\documentclass[10pt,twocolumn,letterpaper]{article}

\usepackage[pagenumbers]{cvpr} 
\usepackage{bm}
\usepackage{multirow}
\usepackage{booktabs}
\usepackage{graphicx}
\usepackage{caption}

\definecolor{cvprblue}{rgb}{0.21,0.49,0.74}
\usepackage[pagebackref,breaklinks,colorlinks,allcolors=cvprblue]{hyperref}

\def\paperID{15816} 
\def\confName{CVPR}
\def\confYear{2026}

\title{SCAIL: Towards Studio-Grade Character Animation via In-Context Learning of 3D-Consistent Pose Representations}

\author{
Wenhao Yan$^{1\ast\dagger}$ \quad
Sheng Ye$^{1\ast\dagger}$ \quad
Zhuoyi Yang$^{1\dagger\ddagger}$ \quad
Jiayan Teng$^{1\dagger}$ \quad
ZhenHui Dong$^{1}$ \\
Kairui Wen$^{1}$ \quad
Xiaotao Gu$^{2}$ \quad
Yong-Jin Liu$^{1\mathsection}$ \quad
Jie Tang$^{1\mathsection}$ \\
\textsuperscript{1}Tsinghua University \quad
\textsuperscript{2}Z.ai
}

\begin{document}

\twocolumn[{
\renewcommand\twocolumn[1][]{#1}
\maketitle

\vspace{-1.0cm}
\begin{center}
\captionsetup{type=figure}
\includegraphics[width=0.95\textwidth]{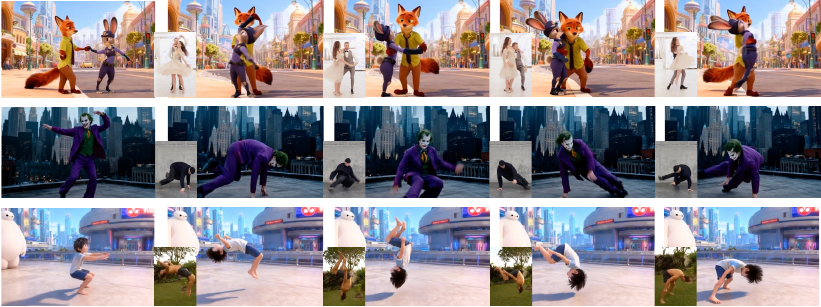}
\vspace{-0.1cm}
\caption{
   We propose \textbf{SCAIL}, a character animation framework that enables high-fidelity character animation under diverse and challenging conditions, including large motion variations, stylized characters, and multi-character interactions.
}
\label{fig:teaser}
\end{center}
\vspace{0.1cm}
}]

\begingroup
\renewcommand\thefootnote{\fnsymbol{footnote}}
\setcounter{footnote}{0}
\footnotetext[1]{Equal contribution.}
\footnotetext[2]{Work done during internship at Z.ai.}
\footnotetext[3]{Project leader.}
\footnotetext[4]{Corresponding author: \texttt{jietang@tsinghua.edu.cn}, \\
\texttt{liuyongjin@tsinghua.edu.cn}}
\endgroup

\begin{abstract}
Achieving controllable character animation that meets studio-grade standards remains challenging despite recent progress. Existing approaches can transfer motion from a driving video to a reference image, but often fail to preserve structural fidelity and temporal consistency in wild scenarios involving complex motion and cross-identity animations. In this work, we present \textbf{SCAIL} (a framework toward \textbf{S}tudio-grade \textbf{C}haracter \textbf{A}nimation via \textbf{I}n-context \textbf{L}earning), which is designed to address these challenges from two key innovations. First, we propose a novel 3D pose representation, providing a robust and flexible motion signal. Second, we introduce a full-context pose injection mechanism within a diffusion-transformer, enabling effective spatio-temporal reasoning over full motion sequences. To align with studio-grade requirements, we develop a curated data pipeline ensuring both diversity and quality, and establish a comprehensive benchmark for systematic evaluation. Experiments show that \textbf{SCAIL} achieves state-of-the-art performance and advances character animation toward studio-grade controlling. Code and model are available at \href{https://github.com/zai-org/SCAIL}{zai-org/SCAIL}.
\end{abstract}    

\section{Introduction}
\label{sec:intro}

High-fidelity character animation has tremendous potential for film production. Conventional filmmaking pipelines rely on a complicated workflow—involving motion capture, rigging, rendering—that demands expensive hardware and significant expert labor.
Recently, the emergence of video generation models~\cite{VDM, SVD, cogvideox, Wan} introduces a new paradigm: given a reference image and a driving video, such models can directly synthesize animations that follow the target motion in driving video. Similar to real-world production pipelines, existing methods~\cite{MagicAnimate, AnimateAnyone, champ, MimicMotion, Stableanimator, animate-x, Unianimate-dit, Wan-Animate} typically begin by extracting skeletal motion sequences from the driving video as a form of “motion capture”, and then inject this information into a video generation model to perform “rigging” and “rendering”. 
However, these methods often struggle with challenging scenarios, such as complex motions (\textit{e.g.,} turning, rolling, flipping), multi-person interactions (\textit{e.g.,} dancing, hugging, fighting), and cross-domain animation where the reference and driving subjects differ significantly in appearance or body shape, exhibiting distorted appearance or implausible limb occlusions. In this work, we identify these challenges as the primary bottleneck to achieving studio-grade controllable animation. These limitations reveal that current skeletal pose representations and pose-driving generation methods fail to adequately capture 3D structure of driving motions, inter-character spatial and occlusion relationships, and temporal correlations of motion sequences.

First, for motion representation, prior works~\cite{AnimateAnyone, MimicMotion, Stableanimator, Unianimate-dit, VACE, Wan-Animate} typically rely on 2D keypoints (extracted from DWPose~\cite{DWPose}, ViTPose~\cite{ViTPose}, \textit{etc.}), which suffer from noisy predictions and cannot encode occlusion. Some works~\cite{champ, Vividpose} adopt SMPL~\cite{SMPL} mesh controls, which offer strong 3D human priors but cause severe identity leakage. Second, regarding pose injection, a common approach in diffusion transformer (DiT)~\cite{DiT} model is to concatenate conditions channel-wise~\cite{realis-dance, Wan-Animate}. However, we find that this method provides more local motion cues than capturing global motion dependencies. 

To address the above limitations, we present \textbf{SCAIL}, a framework that revisits two core technical bottlenecks: (1) how to build a pose representation that effectively bridges driving and generated videos with \textbf{unambiguous and accurate motion encoding}, and (2) how to inject pose control in a way that enables the model to \textbf{capture spatio-temporal motion structures}. We propose a novel 3D pose representation that rasterizes bones as spatial cylinders into the pixel plane, such representation can be augmented and retargeted to enable seamless controlling across diverse characters and scenarios. To prevent model from following local motion cues as a shortcut, we propose a full-context pose injection mechanism within a DiT~\cite{DiT}-based architecture. This design allows the model to attend to the entire pose sequence when generating each frame, enabling it to reason about motion context across time, and better capture high-level motion semantics.

To obtain reliable video-pose pairs, we apply a segmentation-and-extraction approach to get 3D keypoints in human interactions. Building on that, a data curation pipeline ensuring both character diversity and motion complexity is established, with automatic filtering based on human presence and motion-based metrics and a final manual stage to select a finetuning set of superior quality. We also observe that current evaluations lack a comprehensive benchmark that adequately reflects production-level requirements. To address this gap, we propose \textbf{Studio-Bench}. It consists of two parts: the first evaluates motion adherence and structural integrity under complex single- and multi-person actions, while the second measures model generalization when the reference image and driving video differ in identity or domain. Our \textbf{Studio-Bench} covers challenging real-world cases and provides a realistic and rigorous measure of studio-level generation capability.

Our main contributions can be summarized as follows: 
(1) We propose an identity-agnostic 3D pose representation that serves as motion-driving signal suitable for complex motions and multi-human interactions.
(2) We inject driving-pose controls via in-context reasoning to enable effective spatiotemporal motion modeling, yielding superior results in complex and multi-person scenarios.
(3) We construct a pipeline for curating high-quality, diverse training data, and establish a comprehensive \textbf{Studio-Bench} for systematic evaluation.
(4) Our \textbf{SCAIL} framework achieves state-of-the-art performance over existing baselines, and advances character image animation toward production-level readiness.

\section{Related Work}
\label{sec:related}

\begin{figure*}
\centering
\includegraphics[width=0.9\linewidth]{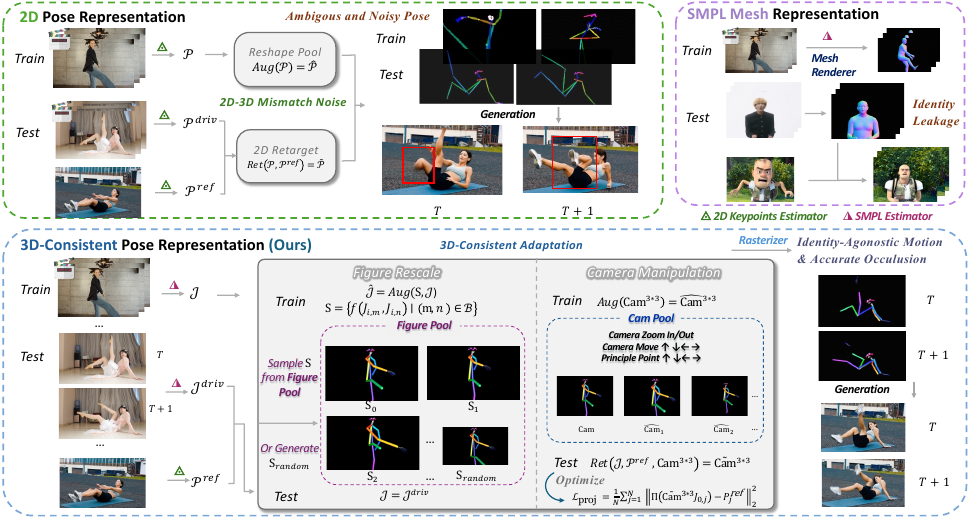}
\caption{Overview of the proposed 3D-consistent pose. For scaling implementation, we take the clavicle or the pelvis as the central reference, applying scaling from proximal to distal along each limb in bones set $\mathcal{B}$. $\textit{Aug}(\cdot)$ denotes augmentation in training, $\textit{Ret}(\cdot)$ denotes retargeting in inference, and \(\mathcal{P}^{\text{ref}} = \{ \text{P}^{\text{ref}}_j \mid 1 \leq j \leq N \}\) denotes $N$ estimated 2D keypoints in the reference image.
We further incorporate hand and face controls by overlaying 2D hand and face keypoints onto the rendered sequences, and align them with the projection of 3D joints during augmentation or retargeting. For better clarity, we omit the drawing process of 2D hand and face in the figure.}
\label{fig:pose}
\vspace{-0.2cm}
\end{figure*}

\subsection{Diffusion Models for Video Generation}
Diffusion models~\cite{ddpm, ddim} effectively overcome the training instability and mode collapse of generative adversarial networks (GANs)~\cite{gan}, and emerge as the dominant paradigm in visual content generation. The success of the Stable Diffusion (SD)~\cite{SD, SDXL} in image synthesis has naturally inspired the extension from image to video.
Subsequently, the Diffusion Transformer (DiT)~\cite{DiT} architecture, combined with RoPE~\cite{rope} for position encoding, offers superior modeling capability and scalability, becoming the leading backbone for high-quality video generation~\cite{sora, kling, cogvideox, Wan}.
More recently, diffusion-based video generation has advanced toward controllability, enabling control over camera viewpoints~\cite{cameractrl, camco}, motion trajectories~\cite{Tora, 3DTrajMaster}, and scene structures~\cite{Gen1, make-your-video, DrivingDiffusion}.
In this work, we focus on character video generation with precise pose control.

\subsection{Controllable Character Image Animation}
Character image animation aims to generate photorealistic and temporally coherent video, where the appearance remains consistent with a given reference image and the motion follows a driving video.
AnimateAnyone~\cite{AnimateAnyone} extracts 2D skeletons from the driving video as motion guidance, and designs Pose Guider and ReferenceNet modules to control motion and appearance. Subsequent works explore various improvements. Champ~\cite{champ} integrates multiple motion signals, including depth maps, SMPL~\cite{SMPL} normals, and 2D skeletons. Animate-X~\cite{animate-x} introduces a skeleton augmentation strategy to enable animation of anthropomorphic characters with large body-ratio discrepancies. Moreover, MimicMotion~\cite{MimicMotion} address hand and facial distortions via regional loss. DanceTogether~\cite{DanceTogether} extends this line of research to multi-person animation.
UniAnimate-DiT~\cite{Unianimate-dit}, VACE~\cite{VACE} and Wan-Animate~\cite{Wan-Animate} replace U-Net with Transformer as the denoising backbone, significantly improving the generation quality. 
Despite progress, current models still struggle with complex scenarios and cross-pair animations, which hinders their deployment in real production.

\section{Method}
\label{sec:method}

\subsection{Preliminaries}
\textbf{Latent Diffusion Models.}
Latent diffusion models~\cite{SD, latent-shift} reduce the computational cost of pixel-space diffusion~\cite{ddpm, Imagen} by operating in a compressed latent space.
Given an input video $\bm{x}$, a pretrained VAE encoder~\cite{Wan} $\mathcal{E}$ first maps it into a latent representation $\bm{z}_0 = \mathcal{E}(\bm{x})$.
During the forward diffusion process, Gaussian noise is progressively added over $T$ timesteps, formulated as:
\begin{equation}
q(\bm{z}_t | \bm{z}_{t-1}) = \mathcal{N}\!\big(\bm{z}_t; \sqrt{1-\beta_t}\,\bm{z}_{t-1},\, \beta_t \mathbf{I}\big),
\end{equation}
where $\beta_t$ denotes the noise schedule.
The denoising model $\bm{\varepsilon}_\theta(\bm{z}_t, t, c)$ learns to predict the added noise conditioned on optional input $c$ (\textit{e.g.,} pose, text), and is optimized by:
\begin{equation}
\mathcal{L} = \mathbb{E}_{\bm{z}_t,\,\bm{\varepsilon} \sim \mathcal{N}(0,\mathbf{I})}\!\big[\|\bm{\varepsilon} - \bm{\varepsilon}_\theta(\bm{z}_t, t, c)\|_2^2\big].
\end{equation}

\noindent\textbf{Diffusion Transformer (DiT).}
Traditional diffusion models typically rely on U-Net as denoising backbones. Recent works propose Diffusion Transformer (DiT)~\cite{DiT} architecture, which supports variable input resolution and sequence length. The DiT combines the generative power of diffusion models with the flexibility of Transformers~\cite{transformer}, enabling better scalability and stronger modeling capacity.



\begin{figure*}
    \centering
    \includegraphics[width=0.93\linewidth]{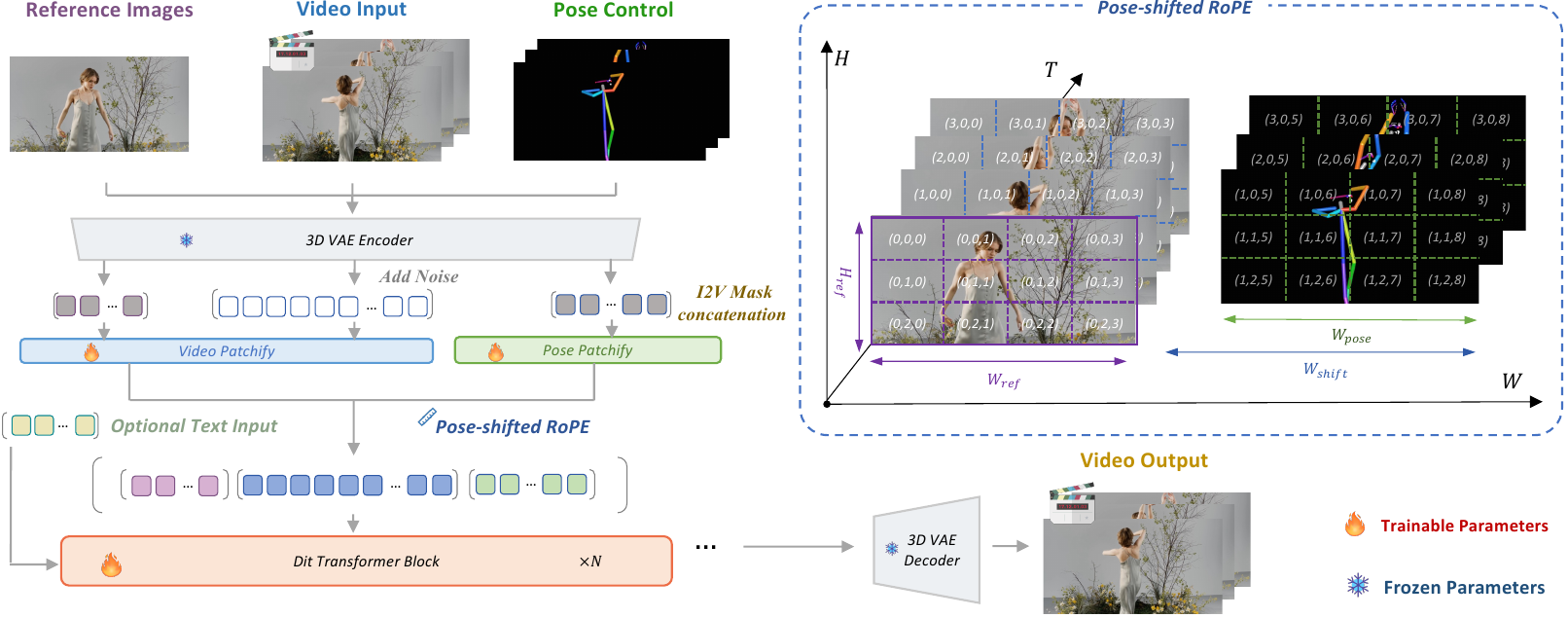}
    \caption{Overview of \textbf{SCAIL}'s model architecture. \textbf{SCAIL} builds upon I2V model and incorporate pose control as an explicit context for the model to learn spatial-temporal motion. To accommodate to the training setting where reference image and video input are sampled from different parts of the video, we modify the I2V model’s input structure by concatenating the reference image at the beginning of the sequence and initiating generation from $T=1$, using the original I2V pattern to inject the reference CLIP feature. To help the model better distinguish the conditional tokens and the noisy video sequence, we leverage the original mask mechanisim of Wan-I2V model architecture, applying an all-one mask for the reference image and the driving sequence, and an all-zero mask for the noisy video sequence. }
    \label{fig:network}
    \vspace{-0.2cm}
\end{figure*}

\subsection{3D-Consistent Pose Conditioning}
\textbf{3D-Consistent Pose Representation.}
Recent methods~\cite{AnimateAnyone, animate-x, Unianimate-dit, VACE, Wan-Animate} that employ 2D keypoint-based skeletons as motion controls perform well on simple actions. However, they often struggle under complex studio-level production scenarios, leading to abnormal human kinematics. As there exists discrepancy between the reference image and the driving pose (e.g. body shape variations), previous 2D-based methods~\cite{animate-x, Unianimate-dit} mitigate such gap through 2D skeleton scaling during training or heuristic retargeting at test time. However, such adaptation inherently suffers from deformation due to inconsistency with 3D motion dynamics, further amplifying estimation noise and inaccuracies.

We design our 3D pose representation to retain depth and occlusion, deliver accurate motion and remain identity-agnostic. To achieve that, we employ NLFPose~\cite{nlfpose} to estimate 3D body keypoints $\mathcal{J} = \{ \text{J}_{i,j} \}_{i=1, j=1}^{T, N}$, where $\tilde{\text{J}}_{i,j}$ denotes the predicted 3D joint $j$ at frame $i$, and connect them according to the skeletal topology in 3D space. We discard SMPL mesh-like representations, which are inherently human-specific and offer limited controllability for augmentation or retargeting, and render 3D skeletons as cylindrical segments to provide driving signals. We distinguish person identities by directly rendering skeletons with different hues instead of relying on hardcoded network designs~\cite{DanceTogether}.

To address the disparity between the reference image and the driving pose, we introduce a two-stage optimization strategy leveraging the 3D prior of our representation. Unlike previous methods~\cite{animate-x} that applies 2D scaling in training augmentation and non-robust skeleton scaling rules in inference retargeting, we design training pose to be fully identity agnostic with accurate bone angles, and in inference we avoid altering the original estimated skeleton and only optimize the camera instead. To achieve identity decoupling during training, we synthesize a set of scale parameters $\mathcal{S}$, and randomly sample $S_i \in \mathcal{S}$ during training, each $S_i$ denoting a set of scaling parameter of each bones, simulating binding motion to characters with diverse body shapes. Since the underlying estimated skeleton topology and relative bone directions remain consistent, the scaled skeletons is temporally coherent and still faithfully represent the same motions. 

In inference, we perform spatial alignment using a projection loss ($\mathcal{L}_{\text{proj}}$) to remap the predicted 3D driving poses to the reference 2D frame to help the model anchor where the motion should be transferred to, and optimize the training process to 
Specifically, we optimize the following objective to refine the camera projection matrix $\tilde{\text{Cam}}^{3\times3}$:
\begin{equation}
\mathcal{L}_{\text{proj}} = \frac{1}{N} \sum_{j=1}^{N} 
\left\| 
\Pi\!\left(\tilde{\text{Cam}}^{3\times3} \tilde{\text{J}}_{0,j}\right) 
- \text{P}^{\text{ref}}_j 
\right\|_2^2,
\end{equation}
where $\Pi(\cdot)$ is the perspective projection, $N$ is joint number, and $\text{P}^{\text{ref}}_j$ is the estimated 2D joints of the reference frame.

As the alignment stage in inference may introduces approximation error, we simulates such error in training by directly applying such aligning method using the reference image and the first frame of the driving sequence (the reference image and driving sequence are sampled from different parts of the video) and exerting modest disturbance on camera parameters ${\text{Cam}}^{3\times3}$ to enhance the model's robustness towards camera variance. Such design strikes a balance between the controllability of location and robust motion transfer.

\

\begin{figure}[t]
    \centering
    \includegraphics[width=0.9\linewidth]{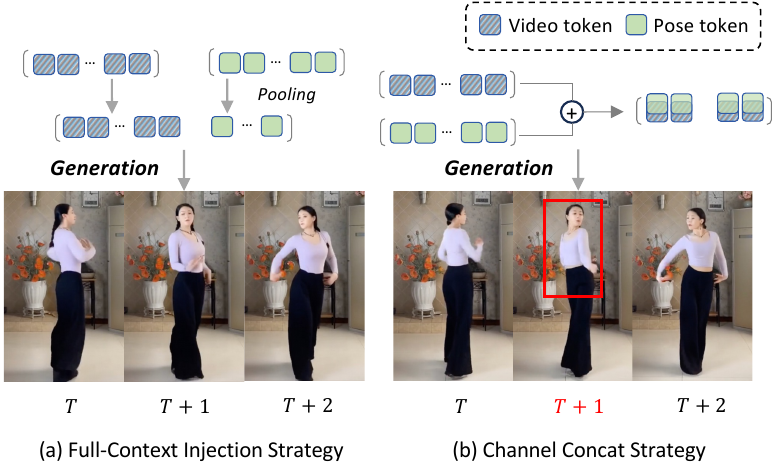}
    \caption{Exploration of different strategies for pose injection.}
    \label{fig:netcompare}
    \vspace{-0.2cm}
\end{figure}

\noindent\textbf{Full-Context Driving Pose Injection.}
Our model builds upon a DiT-based Image-to-Video (I2V) model~\cite{Wan} and inject pose control signals for motion guidance. The common practice for pose information injection is through channel concat~\cite{realis-dance, Wan-Animate} or using a pose adapter~\cite{Unianimate-dit}, where the driving pose sequence is embedded and added to the noisy video latents. We first implement the channel concat approach as illustrated in Figure~\ref{fig:netcompare}, following~\cite{realis-dance}. While this method shows decent pose-following capability in generation, we find it tend to generate unnatural human pose when the motion is complex despite conditioned on accurate and unambiguous pose. A typical example is the turning motion, where the model often fails to correctly distinguish between front and back views, generating awkward postures. We assume that this limitation stems from the per-frame addition scheme, which fails to provide sufficient temporal context for motion understanding. Since motion is inherently sequential, many actions can only be interpreted correctly within a temporal context. Therefore, we design \textit{full-context pose injection}, which enables the model to reason over the entire pose sequence. Specifically, we directly concatenates conditional pose tokens with noisy video tokens to facilitate spatial-temporal interactions between the two modality. As shown in Figure~\ref{fig:netcompare}, this strategy handles motion correctly in complex scenarios.

To mitigate the sequence length increase introduced by our context-aware injection scheme, we apply spatial downsampling to the pose video. Empirically, we find that with 2× downsampling the pose following ability is nearly unaffected. Thus, we adopt this as the default setting in both training and testing, achieving balance between generation quality and efficiency.

\ 

\noindent\textbf{Shifted RoPE for Pose Context.}
Conventional pipelines typically extract poses that are spatially aligned with the original video. However, in our setting, augmentations such as scaling and camera transformations often prevent the extracted pose context to be spatially aligned with the driving video. To address this misalignment, we introduce \textit{Pose-Shifted RoPE} mechanism, which enables the model to effectively retrieve driving signals from the augmented full-sequence pose representation. 

As illustrated in Figure~\ref{fig:network}, conventional 3D RoPE encodes position in the form $(t, h, w)$, where $t$ is the temporal frame index and $h, w$ denote spatial dimensions, each starting from zero. In our design, \textit{Pose-Shifted RoPE} applies a shift along the width dimension specifically to pose representations. The positional vectors for each driving pose token $d$ are defined as:
\begin{equation}
    \text{Pos} = [t,\, h,\, W_{\max} : W_{\max} + \text{shift}_W],
\end{equation}
where $\text{shift}_W$ is a constant shift magnitude and $W_{\max}$ is the width of the reference image. As show in Figure~\ref{fig:network}, this design helps the model to distinguish driving pose tokens from reference image tokens and noisy video tokens in conjunction with the modified I2V Mask and reference image token injection strategy, enhancing the overall performance. Considering the modality gap between the noisy video tokens and the driving pose tokens, we observe that the model performs best particularly when the shift distance is a relatively large constant. To accommodate different downsampling ratios for pose context, mean-pooling on the 3D-RoPE~\cite{rope} frequencies is performed according to the applied ratio.

\begin{figure*}
    \centering
    \includegraphics[width=0.93\linewidth]{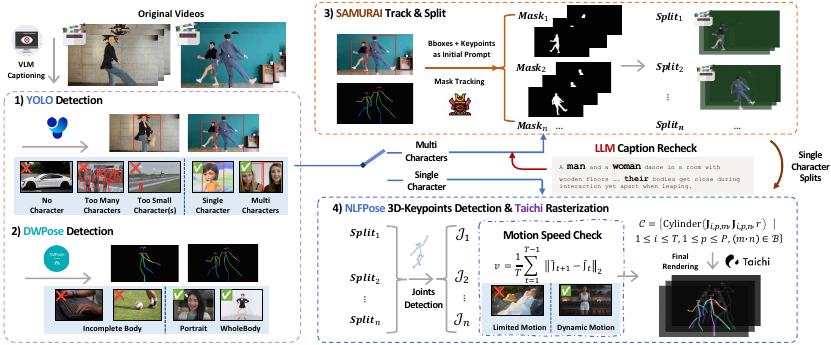}
    \caption{The data curation pipeline. We perform character filtering and motion-speed filtering to construct high-quality training data.}
    \label{fig:data}
    \vspace{-0.2cm}
\end{figure*}

\subsection{Data Pipeline}
\noindent\textbf{Pose Extraction.}
We adopt ~\cite{nlfpose} for 3D keypoints extraction. However, the default extraction pipeline often fails to detect correct limbs under occlusion, especially in multi-human interactions. To address so, we apply a segmentation-and-extraction approach to obtain reliable poses in human interactions. To be more specific, we employ Samurai~\cite{samurai} to track and split the mask of each character, generating multiple single-human video splits. Subsequently, 3D keypoints are extracted for each video that contain one main character. Finally, The 3D skeletons are represented as cylindrical segments in 3D space. Those cylindrical segments are then composed together in 3D space and rasterized on the 2D canvas. This procedure preserves inter-person occlusion relationships, benefiting from NLFPose’s accurate depth estimation. In practice, we find that our proposed multi-stage pipeline provides more accurate estimation than direct multi-person pose estimation methods like PromptHMR~\cite{PromptHMR}, especially in wild cases involving complex interaction. We follow the same extraction pipeline in inference.

\noindent\textbf{Data Curation.}
To meet the requirements for generating studio-grade character animation, we have established the following criteria for data filtering: 
(1) the character should be the primary focus of each frame,  
(2) the person should exhibit explicit motion, and
(3) the pose should be visually complete, covering either the upper body or the full body.
We first detect human presence using YOLO~\cite{yolo} and discard clips where characters are not the main subjects.
Next, 2D keypoints are extracted using DWPose~\cite{DWPose}, retaining only half-body or full-body videos. After this stage, multi-person videos can be further separated based on the number of bounding boxes detected by YOLO~\cite{yolo}. We further employ a large language model~\cite{glm} to analyze captions, complementing the complex rule-based logic of multi-person motion bounding box checking. To compel the model to learn from complex motion dependencies, we compute the \textit{motion speed} from the estimated 3D keypoints and discard samples with limited motion, resulting in a clean and motion-rich training set.
The \textit{motion speed} $v$ is calculated as:
\begin{equation}
    v=\frac{1}{T} \sum_{t=1}^{T-1} \sum_j\left\|\bar{\mathrm{J}}_{t+1, j}-\bar{\mathrm{J}}_{t, j}\right\|_2
\end{equation}
where j refers to the joints in the screen and $\bar{\text{J}}_t = \text{J}_t - \text{J}_t^{root}$ denotes the 3D human joint positions relative to the body center at frame $t$. 

We collect around 250K high-quality motion-rich video-pose pairs using the filtering pipeline, among which 20K pairs featuring multiple characters. To further enhance model's performance on complex motions, we select 12K samples with the highest \textit{motion speed} from dance, general motion, and obtain a finetuning set of 4,000 high-dynamic videos with minimal blur after manual clarity inspection.

\begin{table*}[t]
\small
\centering
\renewcommand{\arraystretch}{1.1}
\setlength{\tabcolsep}{8pt}
\begin{tabular}{lcccc|cccc}
\toprule
\multirow{2.5}{*}{\textbf{Methods}} &
\multicolumn{4}{c|}{\textbf{Self-Driven Animation}} &
\multicolumn{4}{c}{\textbf{Cross-Driven Animation}} \\
\cmidrule{2-9}
& PSNR$\uparrow$ & SSIM$\uparrow$ & LPIPS$\downarrow$ & FVD$\downarrow$ 
& Mot-Acc$\uparrow$ & Kin-Consis$\uparrow$ & Phy-Consis$\uparrow$ & ID-Sim$\uparrow$ \\
\midrule
UniAnimate-DiT~\cite{Unianimate-dit} & 17.79 & 0.637 & 0.242 & 362.27 & 2.5\% & 1.7\% & 0.8\% & 1.7\% \\
VACE~\cite{VACE} & 16.73 & 0.588 & 0.263 & 264.71 & 9.2\% & 14.2\% & 18.3\% & 32.5\% \\
Wan-Animate~\cite{Wan-Animate} & 18.54 & 0.648 & 0.221 & 187.61 & 35.0\% & 28.3\% & 24.2\% & 20.0\% \\
\midrule
\textbf{SCAIL}-14B (Ours)  & \textbf{19.22} & \textbf{0.660} & \textbf{0.206} & \textbf{176.16}
& \textbf{53.3\%} & \textbf{55.8\%} & \textbf{56.7\%} & \textbf{45.8\%} \\
\bottomrule
\end{tabular}
\caption{Quantitative comparison for \textbf{SCAIL}-14B and baselines. All compared methods are built upon 14B Wan~\cite{Wan} foundation models.}
\label{tab:quant_comp_14B}
\end{table*}

\subsection{Studio-Bench for Evaluation}
To evaluate the model capabilities under studio-grade standards, we propose \textbf{Studio-Bench}, a new evaluation benchmark tailored for Studio-Grade character animation. Previous evaluations~\cite{MagicAnimate, AnimateAnyone, champ, MimicMotion, Stableanimator, Unianimate-dit} primarily focus on simple actions, which fail to capture the challenges present in film production, such as complex and dynamic motions, multi-person interactions, and cross-domain animations. Specifically, our benchmark consists of two parts: \textit{Self-Driven} and \textit{Cross-Driven}.
The first part evaluates motion adherence under complex actions. We examine whether the generated videos maintain correct body structure during large-scale motions, and whether the model can accurately capture inter-person relationships in multi-character scenes, totally containing 130 clips. Quantitative metrics can be computed by directly comparing generated results with paired ground-truth videos for this subset.
The second part measures the model’s transferability when the reference image and driving video differ in identity or domain. We manually select a set of real reference images to test generalization, and further construct additional reference images using an image generation tool~{\cite{seedream}} to introduce variations in style and body shape, obtaining 120 single-character pairs and 10 multi-character pairs. As we simulate in-the-wild reference-driving discrepancy, pose retarget should be enabled for single-character pairs in this subset. We incorporate the retarget logic of Unianimate-dit~\cite{Unianimate-dit} for VACE~\cite{VACE} which do not natively support retargeting. For this subset, we conduct user study to assess the generation quality. All quantitative results on the subset are based on the 120 single-character pairs to ensure fairness considering some baselines~\cite{viggle} are incompatible with multi-character settings. Evaluation samples are strictly excluded from the training data.
\section{Experiments}
\label{sec:experiment}

\subsection{Implementation Details}
We train two versions (1.3B and 14B) of our model.
The 1.3B model is finetuned from the Wan2.1-1.3B-Fun-Inp backbone on our pretraining dataset for 6{,}000 steps with a batch size of 96 and a learning rate of 1e-5, using 32 NVIDIA H100 GPUs for approximately two days.
For the larger 14B model, we finetune it from the Wan2.1-I2V-14B backbone in two stages: during the pretraining stage, we train for 8{,}000 steps with a batch size of 96 at a learning rate of 1e-5; after convergence, we perform an additional finetuning stage for 400 steps with the same batch size and a reduced learning rate of 4e-6. Training of the 14B model is conducted on 128 NVIDIA H100 GPUs for over four days with sequence parallelism enabled.
All models are optimized using AdamW~\cite{adamw}. During inference, we set the classifier-free guidance (CFG) scale~\cite{cfg} to 4, offering a favorable balance between pose following and video fidelity.

\begin{figure}[t]
    \centering
    \includegraphics[width=\linewidth]{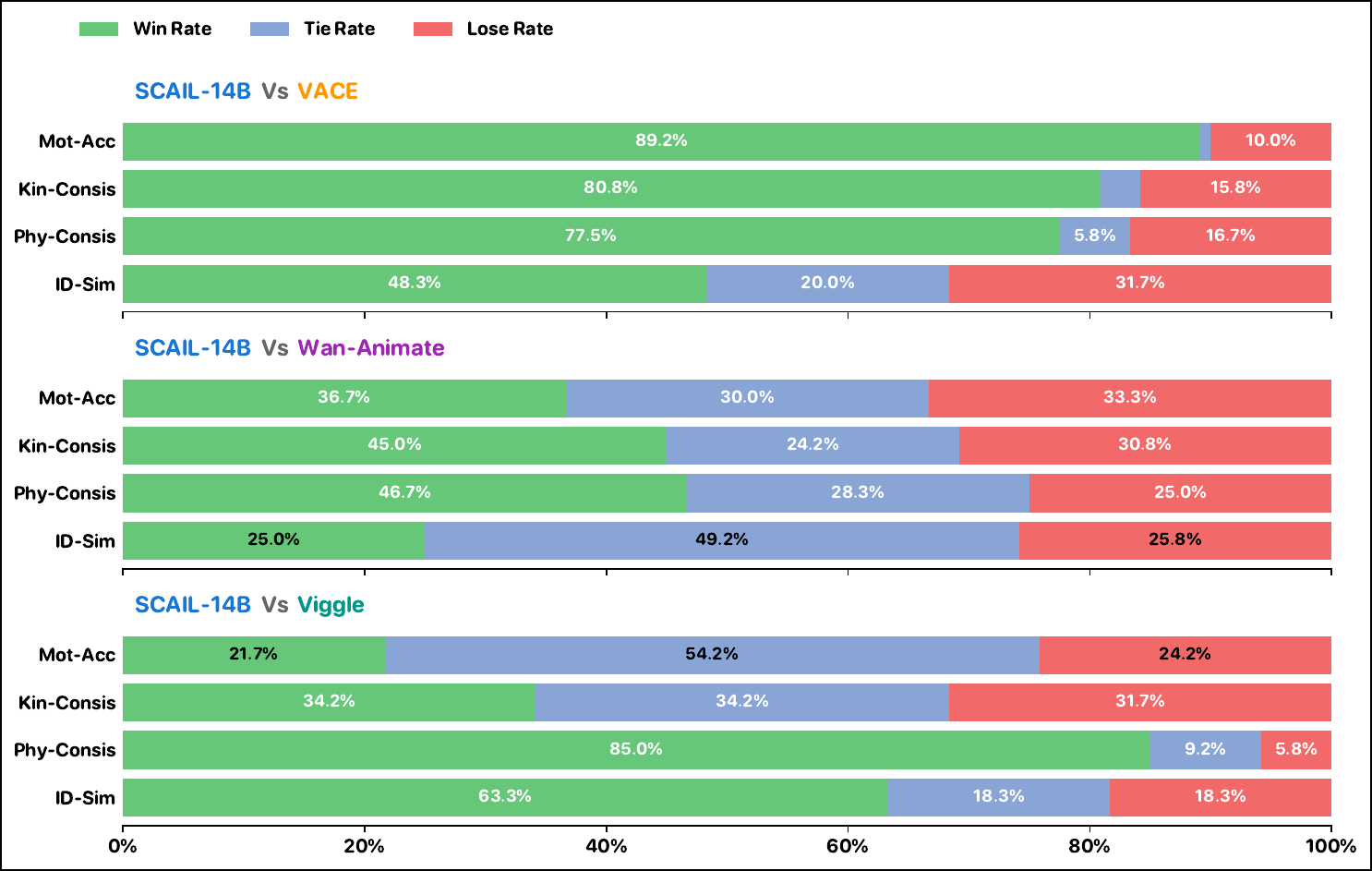}
    \caption{User study for comparing our model with popular community and commercial projects.}
    \label{fig:use study}
\end{figure}

\subsection{Quantitative Evaluation}
We conduct a quantitative comparison with state-of-the-art methods on our proposed \textbf{Studio-Bench}. For \textit{self-driven} part we employed several widely-used quantitative metrics, including PSNR~\cite{psnr}, SSIM~\cite{ssim}, LPIPS~\cite{lpips}, and FVD~\cite{fvd}. To evaluate the generated results in the second subset, we design four metrics: 
(1) \textit{Motion Accuracy}, which measures how faithfully the generated motion follows the driving signal in a frame-by-frame manner.
(2) \textit{Kinesiology Consistency}, evaluating whether joint rotations and body movements remain anatomically feasible and temporally coherent, penalizing sudden twists or physically impossible poses that break natural motion continuity.
(3) \textit{Physical Consistency}, assessing whether the generated motions comply with basic physical constraints such as gravity, support, and momentum conservation, penalizing unrealistic behaviors like hovering in midair.
(4) \textit{Identity Similarity}, measuring the consistency of the subject’s appearance with the reference image.
We convey a detailed blinded user study to collect these metrics, letting users vote for the best in baseline comparison. We also conduct a Win/Tie/Lose study to evaluate our performance between two commonly used open-source frameworks, as well as the closed-source commercial product Viggle~\cite{viggle}, omitting UniAnimate-DiT\cite{Unianimate-dit} due to obvious artifacts under studio-grade demandings.
Viggle is widely believed to rely on a 3D foundation model rather than video diffusion and can be a strong baseline for frame-by-frame motion accuracy under complex scenarios. 

To ensure methodological rigor, we note that the user study for selecting the best-performing model in Table~\ref{tab:quant_comp_14B} and the win/tie/lose evaluation in Figure~\ref{fig:use study} were conducted on different batches of participants, allowing the two sets of results to serve as cross-validations. Results in Table~\ref{tab:quant_comp_14B} show that our model performs better than current open-source works on \textbf{Studio-Bench}. Under the best-model selection scheme, our method shows significant improvements across all metrics, and the win/tie/lose study further highlights its advantages in motion-transfer accuracy over other video diffusion methods and state-of-the-art motion naturalness in the metrics of \textit{Kinesiology Consistency} and \textit{Physical Consistency}.

\begin{figure*}
\centering
\includegraphics[width=0.95\linewidth]{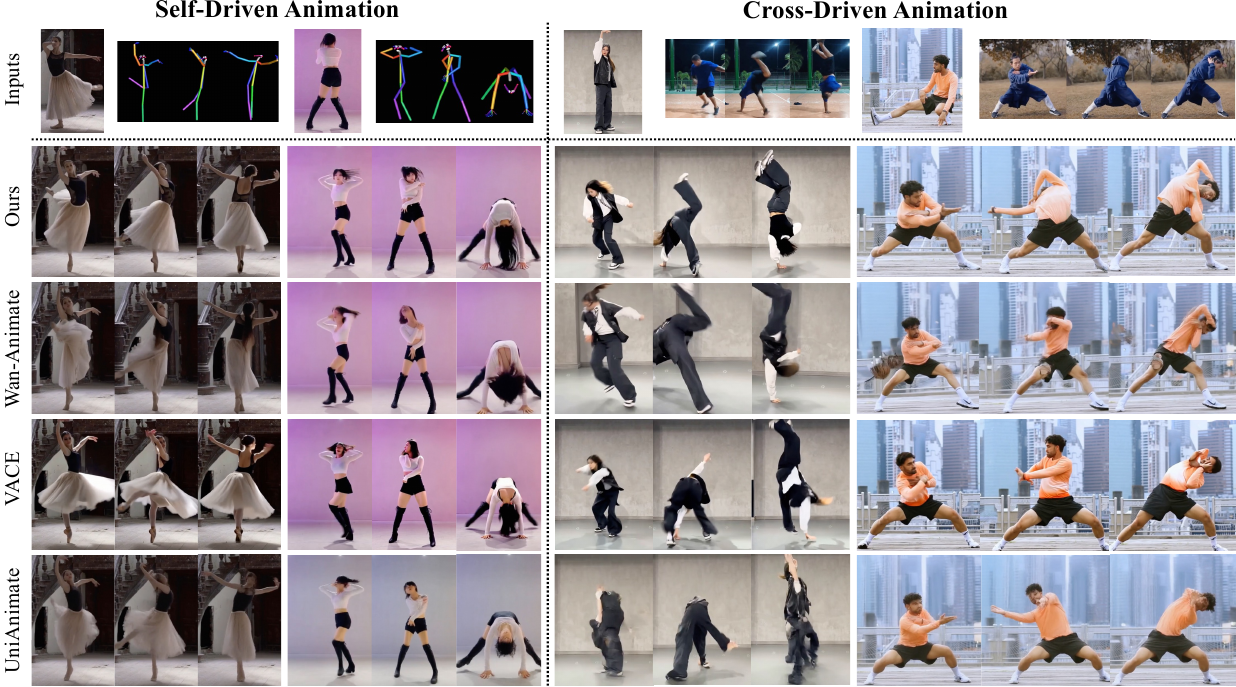}
\caption{Qualitative comparison for single-character animation. Rendered pose in Cross-Driven Animation are omitted for clarity. Zoom in for better visualization.}
\label{fig:qual_comp}
\end{figure*}

\begin{figure*}
\centering
\includegraphics[width=1.0\linewidth]{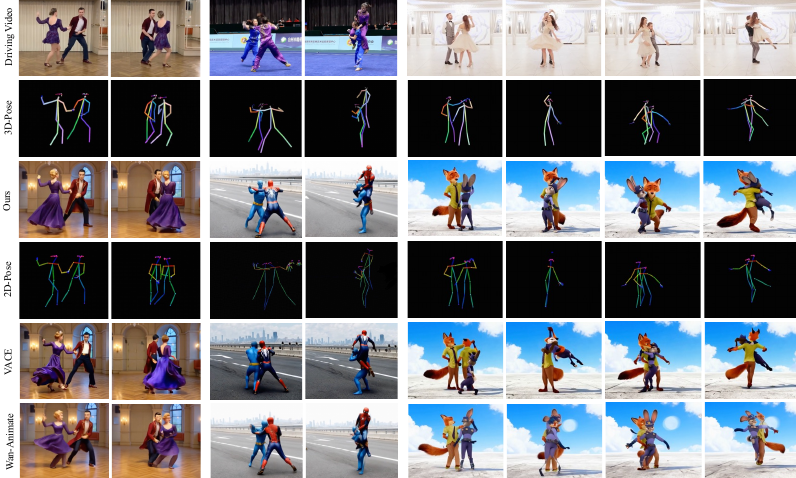}
\caption{Qualitative comparison for dual-character animation. Zoom in for better visualization.}
\label{fig:qual_comp2}
\end{figure*}

\begin{figure}[t]
    \centering
    \includegraphics[width=\linewidth]{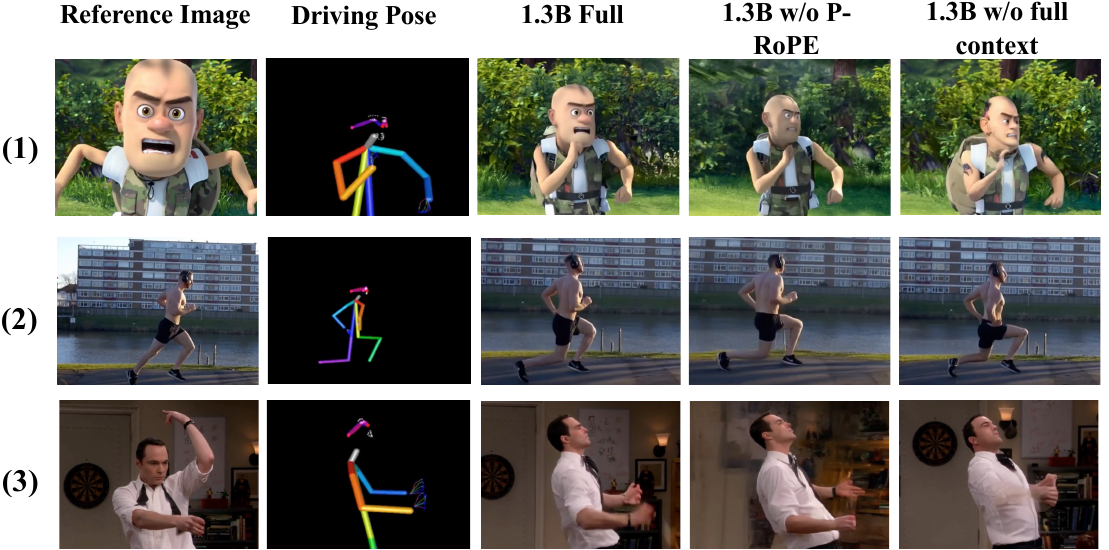}
    \caption{Visualization of ablation study. The \textit{w/o full-context} denotes using the channel-concatenation strategy.}
    \label{fig:ablation}
    \vspace{-0.2cm}
\end{figure}

\subsection{Qualitative Evaluation}
Figure~\ref{fig:qual_comp} shows qualitative comparisons across both self-driven and cross-driven animation settings for single-character animation.
In self-driven cases, \textbf{SCAIL} produces animations with stable structure and accurate limb articulation, particularly in challenging motions such as spinning and bending. Baseline methods often exhibit incomplete pose following or limb abnormalities such as incorrect legs in ballet dancing. In cross-driven cases, our method still demonstrates stronger motion transferability even when large discrepancies exist between the character image and the first frame of the driving video. In contrast, prior methods produce artifacts such as body-shape drift, inconsistent textures, and structural distortions like arm–body interpenetration during acrobats. 

Figure~\ref{fig:qual_comp2} presents qualitative comparisons for multi-character animations, where issues caused by occlusion and estimation error become more pronounced. In relatively clean cases without severe occlusions—such as the dancing scenario in the castle—baseline methods mainly struggle with identity and orientation ambiguities during character turning, similar to their failure patterns in single-character animation. However, as complexity increases, as illustrated by the last two cases of Figure~\ref{fig:qual_comp2}, pose overlapping makes it difficult for the baselines to correctly distinguish human limbs, causing the models to generate severe artifacts where body parts of different characters merge during motion. Furthermore, in multi-character settings, both 2D and 3D pose estimations may occasionally fail due to heavy occlusions (in the shown case they detect only the front character in certain frame), posing significant challenges for temporal and spatial reasoning. In contrast to the baselines which are unable to infer coherent actual poses in such scenarios, our model captures global pose-identity relationships and generates plausible results. These results highlight \textbf{SCAIL}’s strong ability to handle both large motion variations and cross-domain appearance gaps, producing videos that are more natural and visually appealing.

\subsection{Ablation Study}
\label{sec:ablation}
We conduct ablation studies on \textbf{SCAIL}-1.3B model to evaluate the contribution of each component. For reference, we also compare against the U-Net based MimicMotion~\cite{MimicMotion}, highlighting the advantage of our DiT-based architecture. More ablations would be provided in Suppl. \ref{sec:supp_ablation}.

\noindent\textbf{Ablation on the Pose Representation.}
To validate the effectiveness of our 3D-consistent pose representation, we compare with a 2D keypoints-based pipeline (extracted using DWPose~\cite{DWPose} and denoted as \textit{w/ 2D Pose}). Implementation details of the 2D keypoints-based baseline and qualitative comparisions are also provided in Suppl.\ref{sec:supp_ablation}. Results shows the pose representation is critical for the performance gain, and such improvements are more pronounced under full-context pose injection than under channel-concat scheme.

\noindent\textbf{Ablation on Full-Context Driving Pose Injection.}
As shown in Table~\ref{tab:ablation}, replacing channel concatenation with the proposed full-context injection consistently improves all the metrics when providing the model with temporally coherent and information-rich conditions.
Qualitatively, We observe that generated videos exhibit fewer artifacts like limb tearing and structural collapse (shown in Figure~\ref{fig:ablation}), indicating that our context-aware injection strategy enables the model to capture global motion dependencies and reason plausible human poses. Specifically, case (2) in ~\ref{fig:ablation} shows that through this strategy, when the estimator misidentifies the left part and the right part and extracts a posture like a forward lunge in a running driving video, our model with full-context pose injection can still generate correct running posture based on the global motion, as conditions from the reference frame and estimations in other frames clearly reflect the semantic context of running along the river.

\begin{table}
\small
\centering
\renewcommand{\arraystretch}{1.1}
\setlength{\tabcolsep}{2.5pt}
\begin{tabular}{lcccc}
\toprule
\multirow{2.5}{*}{\textbf{Methods}} &
\multicolumn{4}{c}{\textbf{Self-Driven Animation}} \\
\cmidrule{2-5}
& PSNR$\uparrow$ & SSIM$\uparrow$ & LPIPS$\downarrow$ & FVD$\downarrow$ \\
\midrule
\textbf{SCAIL}-1.3B (Ours)  & \textbf{18.08} & \textbf{0.639} & \textbf{0.249} & \textbf{228.62} \\
\textbf{SCAIL}-1.3B \textit{w/ 2D Pose} & 17.08 & 0.619 & 0.284 & 295.36 \\
\textbf{SCAIL}-1.3B \textit{w/o P-RoPE} & 17.79 & 0.637 & 0.280 & 269.35 \\
\textbf{Channel Concat}-1.3B & 17.69 & 0.626 & 0.262 & 263.63 \\
\textbf{Channel Concat}-1.3B \textit{w/ 2D Pose} & 17.12 & 0.624 & 0.282 & 296.23 \\
MimicMotion~\cite{MimicMotion} & 17.01 & 0.630 & 0.314 & 334.24 \\
\bottomrule
\end{tabular}
\caption{Ablation study conducted on \textbf{SCAIL}-1.3B.}
\label{tab:ablation}
\vspace{-0.2cm}
\end{table}

\ 

\noindent\textbf{Ablation on Pose-Shifted RoPE.}
Removing \textit{Pose-Shifted RoPE} (\textit{P-RoPE}) leads to noticeable performance degradation (Table~\ref{tab:ablation}), especially in LPIPS and FVD, confirming that precise pose-aware positional encoding is crucial for maintaining image quality and temporal motion smoothness. Figure~\ref{fig:ablation} shows that \textit{P-RoPE} can strengthen the correspondence between motion cues and spatial structure, resulting in more accurate hand articulation and more reasonable foot grounding. Results from case (1) demonstrate that with full-context injection, \textit{P-RoPE} yields the strongest disentanglement of character identity and pose guidance, enabling the model to faithfully preserve the subject’s appearance while following motion patterns.

\section{Conclusion}

In this work, we present \textbf{SCAIL}, a novel framework for studio-grade character image animation. By introducing a scalable and robust 3D pose representation and leveraging a novel full-context pose injection with shifted RoPE, we enhances spatiotemporal reasoning within a DiT architecture, allowing the model to generate structurally accurate and temporally consistent animations under challenging scenarios. With curated data pipeline and comprehensive Studio-Bench, we push character animation towards production-level standards. Experiments show that SCAIL achieves state-of-the-art performance in both self-driven and cross-driven animation. We believe this work provides a solid step toward practical, production-ready character animation.
\clearpage
{
    \small
    \bibliographystyle{ieeenat_fullname}
    \bibliography{main}

@String(CVPR= {IEEE Conf. Comput. Vis. Pattern Recog.})

@String(ICLR = {Int. Conf. Learn. Represent.})

@String(CVPR  = {CVPR})

@String(ICLR  = {ICLR})

@inproceedings{cogvideox,
  author       = {Zhuoyi Yang and
                  Jiayan Teng and
                  Wendi Zheng and
                  Ming Ding and
                  Shiyu Huang and
                  Jiazheng Xu and
                  Yuanming Yang and
                  Wenyi Hong and
                  Xiaohan Zhang and
                  Guanyu Feng and
                  Da Yin and
                  Yuxuan Zhang and
                  Weihan Wang and
                  Yean Cheng and
                  Bin Xu and
                  Xiaotao Gu and
                  Yuxiao Dong and
                  Jie Tang},
  title        = {CogVideoX: Text-to-Video Diffusion Models with An Expert Transformer},
  booktitle    = {The Thirteenth International Conference on Learning Representations,
                  {ICLR}},
  year         = {2025},
}

@misc{kling,
  author       = {{Kuaishou AI Team}},
  title        = {{Kling}},
  year         = {2024},
  howpublished = {\url{https://kling.kuaishou.com/en}},
}

@misc{sora,
  author       = {{OpenAI}},
  title        = {{Sora}},
  year         = {2024},
  howpublished = {\url{https://openai.com/index/sora/}},
}

@misc{viggle,
  author       = {{Viggle AI}},
  title        = {{Viggle}},
  year         = {2024},
  howpublished = {\url{https://viggle.ai/}},
}

@article{Wan,
  title={Wan: Open and advanced large-scale video generative models},
  author={Wan, Team and Wang, Ang and Ai, Baole and Wen, Bin and Mao, Chaojie and Xie, Chen-Wei and Chen, Di and Yu, Feiwu and Zhao, Haiming and Yang, Jianxiao and others},
  journal={arXiv preprint arXiv:2503.20314},
  year={2025}
}

@inproceedings{AnimateAnyone,
  title={Animate anyone: Consistent and controllable image-to-video synthesis for character animation},
  author={Hu, Li},
  booktitle={Proceedings of the IEEE/CVF Conference on Computer Vision and Pattern Recognition},
  pages={8153--8163},
  year={2024}
}

@inproceedings{MimicMotion,
  title={MimicMotion: High-Quality Human Motion Video Generation with Confidence-aware Pose Guidance},
  author={Zhang, Yuang and Gu, Jiaxi and Wang, Li-Wen and Wang, Han and Cheng, Junqi and Zhu, Yuefeng and Zou, Fangyuan},
  booktitle={International Conference on Machine Learning},
  year={2025}
}

@inproceedings{champ,
  title={Champ: Controllable and consistent human image animation with 3d parametric guidance},
  author={Zhu, Shenhao and Chen, Junming Leo and Dai, Zuozhuo and Dong, Zilong and Xu, Yinghui and Cao, Xun and Yao, Yao and Zhu, Hao and Zhu, Siyu},
  booktitle={European Conference on Computer Vision},
  pages={145--162},
  year={2024},
  organization={Springer}
}

@inproceedings{animate-x,
  author       = {Shuai Tan and
                  Biao Gong and
                  Xiang Wang and
                  Shiwei Zhang and
                  Dandan Zheng and
                  Ruobing Zheng and
                  Kecheng Zheng and
                  Jingdong Chen and
                  Ming Yang},
  title        = {Animate-X: Universal Character Image Animation with Enhanced Motion
                  Representation},
  booktitle    = {The Thirteenth International Conference on Learning Representations,
                  {ICLR}},
  year         = {2025},
}

@article{Unianimate-dit,
  title={Unianimate-dit: Human image animation with large-scale video diffusion transformer},
  author={Wang, Xiang and Zhang, Shiwei and Tang, Longxiang and Zhang, Yingya and Gao, Changxin and Wang, Yuehuan and Sang, Nong},
  journal={arXiv preprint arXiv:2504.11289},
  year={2025}
}

@inproceedings{Stableanimator,
  title={Stableanimator: High-quality identity-preserving human image animation},
  author={Tu, Shuyuan and Xing, Zhen and Han, Xintong and Cheng, Zhi-Qi and Dai, Qi and Luo, Chong and Wu, Zuxuan},
  booktitle={Proceedings of the Computer Vision and Pattern Recognition Conference},
  pages={21096--21106},
  year={2025}
}

@article{Wan-Animate,
  title={Wan-Animate: Unified Character Animation and Replacement with Holistic Replication},
  author={Cheng, Gang and Gao, Xin and Hu, Li and Hu, Siqi and Huang, Mingyang and Ji, Chaonan and Li, Ju and Meng, Dechao and Qi, Jinwei and Qiao, Penchong and others},
  journal={arXiv preprint arXiv:2509.14055},
  year={2025}
}

@inproceedings{MagicAnimate,
  title={Magicanimate: Temporally consistent human image animation using diffusion model},
  author={Xu, Zhongcong and Zhang, Jianfeng and Liew, Jun Hao and Yan, Hanshu and Liu, Jia-Wei and Zhang, Chenxu and Feng, Jiashi and Shou, Mike Zheng},
  booktitle={Proceedings of the IEEE/CVF Conference on Computer Vision and Pattern Recognition},
  pages={1481--1490},
  year={2024}
}

@article{ViTPose,
  title={Vitpose: Simple vision transformer baselines for human pose estimation},
  author={Xu, Yufei and Zhang, Jing and Zhang, Qiming and Tao, Dacheng},
  journal={Advances in neural information processing systems},
  volume={35},
  pages={38571--38584},
  year={2022}
}

@inproceedings{DWPose,
  title={Effective whole-body pose estimation with two-stages distillation},
  author={Yang, Zhendong and Zeng, Ailing and Yuan, Chun and Li, Yu},
  booktitle={Proceedings of the IEEE/CVF International Conference on Computer Vision},
  pages={4210--4220},
  year={2023}
}

@article{Vividpose,
  title={Vividpose: Advancing stable video diffusion for realistic human image animation},
  author={Wang, Qilin and Jiang, Zhengkai and Xu, Chengming and Zhang, Jiangning and Wang, Yabiao and Zhang, Xinyi and Cao, Yun and Cao, Weijian and Wang, Chengjie and Fu, Yanwei},
  journal={arXiv preprint arXiv:2405.18156},
  year={2024}
}

@article{SMPL,
  author       = {Matthew Loper and
                  Naureen Mahmood and
                  Javier Romero and
                  Gerard Pons{-}Moll and
                  Michael J. Black},
  title        = {{SMPL:} a skinned multi-person linear model},
  journal      = {{ACM} Trans. Graph.},
  volume       = {34},
  number       = {6},
  pages        = {248:1--248:16},
  year         = {2015},
}

@inproceedings{DiT,
  title={Scalable diffusion models with transformers},
  author={Peebles, William and Xie, Saining},
  booktitle={Proceedings of the IEEE/CVF international conference on computer vision},
  pages={4195--4205},
  year={2023}
}

@article{ddpm,
  title={Denoising diffusion probabilistic models},
  author={Ho, Jonathan and Jain, Ajay and Abbeel, Pieter},
  journal={Advances in neural information processing systems},
  volume={33},
  pages={6840--6851},
  year={2020}
}

@inproceedings{ddim,
  author       = {Jiaming Song and
                  Chenlin Meng and
                  Stefano Ermon},
  title        = {Denoising Diffusion Implicit Models},
  booktitle    = {9th International Conference on Learning Representations, {ICLR}},
  year         = {2021},
}

@article{gan,
  title={Generative adversarial nets},
  author={Goodfellow, Ian J and Pouget-Abadie, Jean and Mirza, Mehdi and Xu, Bing and Warde-Farley, David and Ozair, Sherjil and Courville, Aaron and Bengio, Yoshua},
  journal={Advances in neural information processing systems},
  volume={27},
  year={2014}
}

@inproceedings{SD,
  title={High-resolution image synthesis with latent diffusion models},
  author={Rombach, Robin and Blattmann, Andreas and Lorenz, Dominik and Esser, Patrick and Ommer, Bj{\"o}rn},
  booktitle={Proceedings of the IEEE/CVF conference on computer vision and pattern recognition},
  pages={10684--10695},
  year={2022}
}

@inproceedings{SDXL,
  author       = {Dustin Podell and
                  Zion English and
                  Kyle Lacey and
                  Andreas Blattmann and
                  Tim Dockhorn and
                  Jonas M{\"{u}}ller and
                  Joe Penna and
                  Robin Rombach},
  title        = {{SDXL:} Improving Latent Diffusion Models for High-Resolution Image
                  Synthesis},
  booktitle    = {The Twelfth International Conference on Learning Representations,
                  {ICLR}},
  year         = {2024},
}

@article{SVD,
  title={Stable video diffusion: Scaling latent video diffusion models to large datasets},
  author={Blattmann, Andreas and Dockhorn, Tim and Kulal, Sumith and Mendelevitch, Daniel and Kilian, Maciej and Lorenz, Dominik and Levi, Yam and English, Zion and Voleti, Vikram and Letts, Adam and others},
  journal={arXiv preprint arXiv:2311.15127},
  year={2023}
}

@article{VDM,
  title={Video diffusion models},
  author={Ho, Jonathan and Salimans, Tim and Gritsenko, Alexey and Chan, William and Norouzi, Mohammad and Fleet, David J},
  journal={Advances in neural information processing systems},
  volume={35},
  pages={8633--8646},
  year={2022}
}

@article{rope,
  title={Roformer: Enhanced transformer with rotary position embedding},
  author={Su, Jianlin and Ahmed, Murtadha and Lu, Yu and Pan, Shengfeng and Bo, Wen and Liu, Yunfeng},
  journal={Neurocomputing},
  volume={568},
  pages={127063},
  year={2024},
  publisher={Elsevier}
}

@article{cameractrl,
  title={Cameractrl: Enabling camera control for text-to-video generation},
  author={He, Hao and Xu, Yinghao and Guo, Yuwei and Wetzstein, Gordon and Dai, Bo and Li, Hongsheng and Yang, Ceyuan},
  journal={arXiv preprint arXiv:2404.02101},
  year={2024}
}

@article{camco,
  title={Camco: Camera-controllable 3d-consistent image-to-video generation},
  author={Xu, Dejia and Nie, Weili and Liu, Chao and Liu, Sifei and Kautz, Jan and Wang, Zhangyang and Vahdat, Arash},
  journal={arXiv preprint arXiv:2406.02509},
  year={2024}
}

@inproceedings{Tora,
  title={Tora: Trajectory-oriented diffusion transformer for video generation},
  author={Zhang, Zhenghao and Liao, Junchao and Li, Menghao and Dai, Zuozhuo and Qiu, Bingxue and Zhu, Siyu and Qin, Long and Wang, Weizhi},
  booktitle={Proceedings of the Computer Vision and Pattern Recognition Conference},
  pages={2063--2073},
  year={2025}
}

@inproceedings{3DTrajMaster,
  author       = {Xiao Fu and
                  Xian Liu and
                  Xintao Wang and
                  Sida Peng and
                  Menghan Xia and
                  Xiaoyu Shi and
                  Ziyang Yuan and
                  Pengfei Wan and
                  Di Zhang and
                  Dahua Lin},
  title        = {3DTrajMaster: Mastering 3D Trajectory for Multi-Entity Motion in Video
                  Generation},
  booktitle    = {The Thirteenth International Conference on Learning Representations,
                  {ICLR}},
  year         = {2025},
}

@article{make-your-video,
  title={Make-your-video: Customized video generation using textual and structural guidance},
  author={Xing, Jinbo and Xia, Menghan and Liu, Yuxin and Zhang, Yuechen and Zhang, Yong and He, Yingqing and Liu, Hanyuan and Chen, Haoxin and Cun, Xiaodong and Wang, Xintao and others},
  journal={IEEE Transactions on Visualization and Computer Graphics},
  volume={31},
  number={2},
  pages={1526--1541},
  year={2024},
  publisher={IEEE}
}

@inproceedings{Gen1,
  title={Structure and content-guided video synthesis with diffusion models},
  author={Esser, Patrick and Chiu, Johnathan and Atighehchian, Parmida and Granskog, Jonathan and Germanidis, Anastasis},
  booktitle={Proceedings of the IEEE/CVF international conference on computer vision},
  pages={7346--7356},
  year={2023}
}

@inproceedings{DrivingDiffusion,
  title={DrivingDiffusion: layout-guided multi-view driving scenarios video generation with latent diffusion model},
  author={Li, Xiaofan and Zhang, Yifu and Ye, Xiaoqing},
  booktitle={European Conference on Computer Vision},
  pages={469--485},
  year={2024},
  organization={Springer}
}

@article{DanceTogether,
  title={DanceTogether! Identity-Preserving Multi-Person Interactive Video Generation},
  author={Chen, Junhao and Chen, Mingjin and Xu, Jianjin and Li, Xiang and Dong, Junting and Sun, Mingze and Jiang, Puhua and Li, Hongxiang and Yang, Yuhang and Zhao, Hao and others},
  journal={arXiv preprint arXiv:2505.18078},
  year={2025}
}

@article{VACE,
  title={Vace: All-in-one video creation and editing},
  author={Jiang, Zeyinzi and Han, Zhen and Mao, Chaojie and Zhang, Jingfeng and Pan, Yulin and Liu, Yu},
  journal={arXiv preprint arXiv:2503.07598},
  year={2025}
}

@article{latent-shift,
  title={Latent-shift: Latent diffusion with temporal shift for efficient text-to-video generation},
  author={An, Jie and Zhang, Songyang and Yang, Harry and Gupta, Sonal and Huang, Jia-Bin and Luo, Jiebo and Yin, Xi},
  journal={arXiv preprint arXiv:2304.08477},
  year={2023}
}

@article{Imagen,
  title={Imagen video: High definition video generation with diffusion models},
  author={Ho, Jonathan and Chan, William and Saharia, Chitwan and Whang, Jay and Gao, Ruiqi and Gritsenko, Alexey and Kingma, Diederik P and Poole, Ben and Norouzi, Mohammad and Fleet, David J and others},
  journal={arXiv preprint arXiv:2210.02303},
  year={2022}
}

@article{transformer,
  title={Attention is all you need},
  author={Vaswani, Ashish and Shazeer, Noam and Parmar, Niki and Uszkoreit, Jakob and Jones, Llion and Gomez, Aidan N and Kaiser, {\L}ukasz and Polosukhin, Illia},
  journal={Advances in neural information processing systems},
  volume={30},
  year={2017}
}

@article{nlfpose,
  title={Neural localizer fields for continuous 3d human pose and shape estimation},
  author={S{\'a}r{\'a}ndi, Istv{\'a}n and Pons-Moll, Gerard},
  journal={Advances in Neural Information Processing Systems},
  volume={37},
  pages={140032--140065},
  year={2024}
}

@article{samurai,
  title={Samurai: Adapting segment anything model for zero-shot visual tracking with motion-aware memory},
  author={Yang, Cheng-Yen and Huang, Hsiang-Wei and Chai, Wenhao and Jiang, Zhongyu and Hwang, Jenq-Neng},
  journal={arXiv preprint arXiv:2411.11922},
  year={2024}
}

@article{taichi,
  title={Taichi: An open-source computer graphics library},
  author={Hu, Yuanming},
  journal={arXiv preprint arXiv:1804.09293},
  year={2018}
}

@inproceedings{yolo,
  title={Yolov8: A novel object detection algorithm with enhanced performance and robustness},
  author={Varghese, Rejin and Sambath, M},
  booktitle={2024 International conference on advances in data engineering and intelligent computing systems (ADICS)},
  pages={1--6},
  year={2024},
  organization={IEEE}
}

@article{glm,
  title={Chatglm: A family of large language models from glm-130b to glm-4 all tools},
  author={GLM, Team and Zeng, Aohan and Xu, Bin and Wang, Bowen and Zhang, Chenhui and Yin, Da and Zhang, Dan and Rojas, Diego and Feng, Guanyu and Zhao, Hanlin and others},
  journal={arXiv preprint arXiv:2406.12793},
  year={2024}
}

@article{realis-dance,
  title={RealisDance-DiT: Simple yet Strong Baseline towards Controllable Character Animation in the Wild},
  author={Zhou, Jingkai and Wu, Yifan and Li, Shikai and Wei, Min and Fan, Chao and Chen, Weihua and Jiang, Wei and Wang, Fan},
  journal={arXiv preprint arXiv:2504.14977},
  year={2025}
}

@article{seedream,
  title={Seedream 4.0: Toward next-generation multimodal image generation},
  author={Seedream, Team and Chen, Yunpeng and Gao, Yu and Gong, Lixue and Guo, Meng and Guo, Qiushan and Guo, Zhiyao and Hou, Xiaoxia and Huang, Weilin and Huang, Yixuan and others},
  journal={arXiv preprint arXiv:2509.20427},
  year={2025}
}

@inproceedings{adamw,
  author       = {Ilya Loshchilov and
                  Frank Hutter},
  title        = {Decoupled Weight Decay Regularization},
  booktitle    = {7th International Conference on Learning Representations, {ICLR}},
  year         = {2019}
}

@article{cfg,
  title={Classifier-free diffusion guidance},
  author={Ho, Jonathan and Salimans, Tim},
  journal={arXiv preprint arXiv:2207.12598},
  year={2022}
}

@inproceedings{psnr,
  title={Image quality metrics: PSNR vs. SSIM},
  author={Hore, Alain and Ziou, Djemel},
  booktitle={2010 20th international conference on pattern recognition},
  pages={2366--2369},
  year={2010},
  organization={IEEE}
}

@article{ssim,
  title={Image quality assessment: from error visibility to structural similarity},
  author={Wang, Zhou and Bovik, Alan C and Sheikh, Hamid R and Simoncelli, Eero P},
  journal={IEEE transactions on image processing},
  volume={13},
  number={4},
  pages={600--612},
  year={2004},
  publisher={IEEE}
}

@inproceedings{lpips,
  title={The unreasonable effectiveness of deep features as a perceptual metric},
  author={Zhang, Richard and Isola, Phillip and Efros, Alexei A and Shechtman, Eli and Wang, Oliver},
  booktitle={Proceedings of the IEEE conference on computer vision and pattern recognition},
  pages={586--595},
  year={2018}
}

@article{fvd,
  title={Towards accurate generative models of video: A new metric \& challenges},
  author={Unterthiner, Thomas and Van Steenkiste, Sjoerd and Kurach, Karol and Marinier, Raphael and Michalski, Marcin and Gelly, Sylvain},
  journal={arXiv preprint arXiv:1812.01717},
  year={2018}
}

@InProceedings{PromptHMR,
    author    = {Wang, Yufu and Sun, Yu and Patel, Priyanka and Daniilidis, Kostas and Black, Michael J. and Kocabas, Muhammed},
    title     = {PromptHMR: Promptable Human Mesh Recovery},
    booktitle = {Proceedings of the IEEE/CVF Conference on Computer Vision and Pattern Recognition (CVPR)},
    month     = {June},
    year      = {2025},
    pages     = {1148-1159}
}
}

\clearpage
\setcounter{page}{1}
\maketitlesupplementary

\renewcommand{\thesection}{A\arabic{section}}
\renewcommand{\thesubsection}{A\arabic{section}.\arabic{subsection}}
\renewcommand{\thefigure}{A\arabic{figure}}
\renewcommand{\thetable}{A\arabic{table}}

\setcounter{section}{0}
\setcounter{figure}{0}
\setcounter{table}{0}

\newcommand{\lyj}{\textcolor{red}}

\section{Details on Pose Conditioning}
\label{sec:supp_impl}
\noindent\textbf{Rendering Details.} To preserve spatial relationships between joints, we represent the human skeleton using cylindrical structures. For efficient rendering of multiple skeletons, we employ a 3D rendering pipeline that first converts the cylinders into spatial voxels, followed by ray marching implemented via~\cite{taichi}. This strategy is highly optimized for modern GPUs, introducing negligible computational overhead. To facilitate person discrimination in multi-person motion sequences, we assign distinct color schemes to each individual. This enables the model to directly learn how to distinguish characters from the representation. We observe that this design effectively alleviates identity switching when the positions of the people interchange.

\noindent\textbf{Augmentation Details.} 
As our augmentation strategy is designed to maximally preserve the original motion, we use a high augmentation rate of 0.8 to achieve the balance between the pose-following accuracy and motion transfer robustness. During training, the overlaid 2D hand and face keypoints extracted by DWPose~\cite{DWPose} will go through additional augmenation after the 3D adaptation process where 2D keypoints are shifted to match the reformed 3D skeleton after body rescaling and camera manipulation. This step is to minimize the unintended influence of the 2D facial and hand signals on the 3D pose representation.

\noindent\textbf{Performance Tradeoff.} 
For the injection of the representation in \textit{full-context} settings, we observe that the rendered pose sequence is relatively sparse as most frame areas consist of non-informative black pixels. In our method, \textit{spatial pooling} to pose sequence can serve as a simple workaround to effectively reduce the contextual pose tokens to $1/4$ and preserves accurate pose-following capability. In addition, \textit{full-context pose injection} introduces no new parameters except the additional patchify layer to the original model, offering a more streamlined architecture compared to stacking additional DiT layers in residual context-tuning methods~\cite{VACE}. For quantitative comparison, we compare the inference costs of the two injection schemes (512 * 896 resolution, 81 frames, 20 diffusion steps) on an H100 GPU. In general, considering the significant motion error reduction, we conclude that the modest efficiency trade-off is acceptable, particularly in studio-grade scenarios which prioritize stringent accuracy and stability. 

\begin{table}[!ht] 
\centering
\scriptsize 
\renewcommand{\arraystretch}{0.9} 
\setlength{\tabcolsep}{3pt} 
\vspace{-7pt}

\begin{tabular}{lcccc}
\toprule
\textbf{Methods (all w/ CFG)}  & Inference Time (s) & FPS & Memory (GB) \\
\midrule
14B \textit{w/} channel-concat & 286.11 & 0.283 & 61.7 \\
14B \textit{w/} full-context (Ours) & 380.78 {\tiny (\textbf{+33.1\%})} & 0.213 {\tiny (\textbf{-24.9\%})} & 68.5 {\tiny (\textbf{+11.0\%})} \\[-1pt]
\bottomrule
\end{tabular}
\label{tab:inferencecost}
\vspace{-7pt}
\end{table}

\begin{figure}[t]
    \centering
    \includegraphics[width=\linewidth]{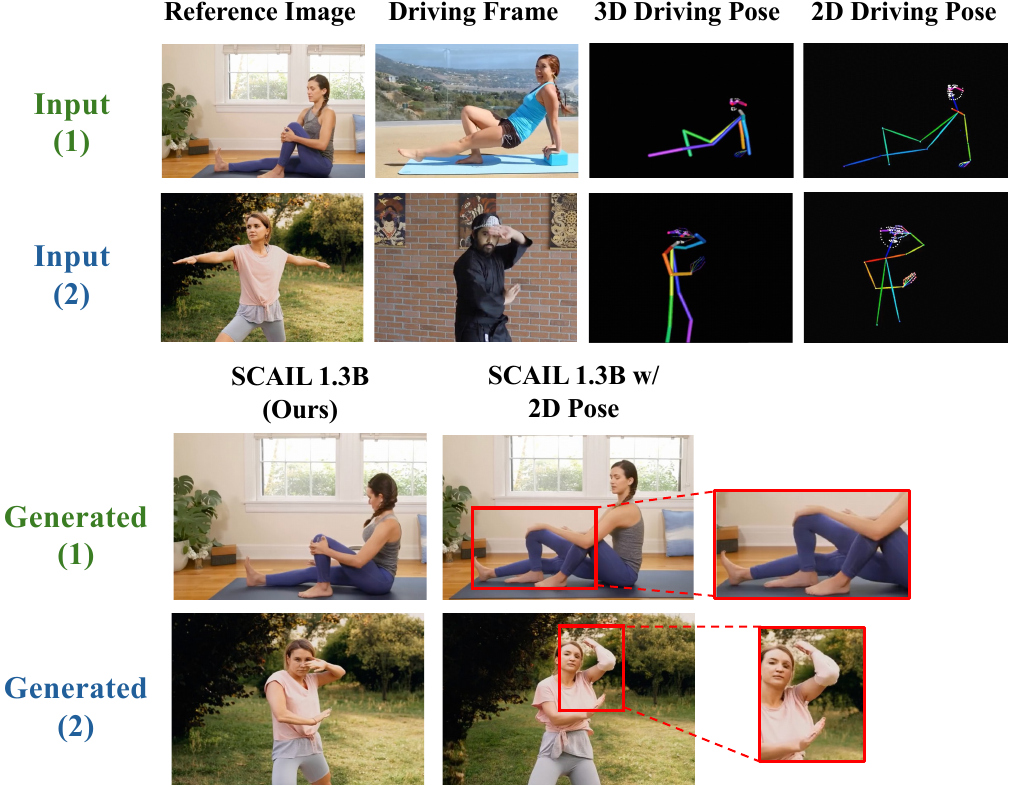}
    \caption{Ablation studies on pose representation. Anomalies in the human body or deviations from correct posture are boxed.}
    \label{fig:ablationpose}
    \vspace{-0.1cm}
\end{figure}

\section{More Ablation Studies}
\label{sec:supp_ablation}
We conduct further ablation studies on \textbf{SCAIL}-1.3B model to evaluate the contribution of our proposed components. All training settings, including learning rate, data, batch size, and training steps, are kept exactly the same across all models in our ablation studies. Additional user studies are conducted to collect users' preference towards different configurations for the \textbf{SCAIL}-1.3B model.

\subsection{Ablation details on the Pose Representation}
As mentioned in the main paper, we implement the 2D augmentation strategy as close as to the 3D version, with similar figure settings and same augmentation ratio of 0.8. Faces and hands are also augmented identically to the 3D version and retargeting logic from~\cite{Unianimate-dit} are applied during inference.  As shown in Table~\ref{tab:ablation}, the estimation noise and the \textit{2D-3D mismatch} noise introduced by the pipeline significantly undermine the model's performance in the self-driven subset of our challenging Studio-Bench, which involves a high ratio of complex motions.

For the cross-driven subset, 2D pose can easily lead to distorted limbs in generation especially when the 1.3B model have difficulty transferring motion to a significant different reference image, as seen in case (1) of Figure~\ref{fig:ablationpose}. Case (2) indicates that when model needs to distinguish the front and back of limbs, the inherent ambiguity of 2D pose can result in incorrect pose interpretations. Qualitative results from the cross-driven subset and quantitative results from the self-driven subset together demonstrate the overall effectiveness of our proposed 3D-based solution. Figure~\ref{fig:ablationret} can also demonstrate the unreasonable scaling factors introduced by 2D-based retargeting process during the inference step, which we will discuss in the next section of retarget ablations.

\subsection{Ablation on 3D-Consistent Adaptation}
Furthermore, we validate the key component in our pose representation: 3D-Consistent Adaptation. 3D-Consistent Adaptation includes 3D Retarget in cross-driven inference and 3D Augmentation in training.

\begin{figure}[t]
    \centering
    \includegraphics[width=0.95\linewidth]{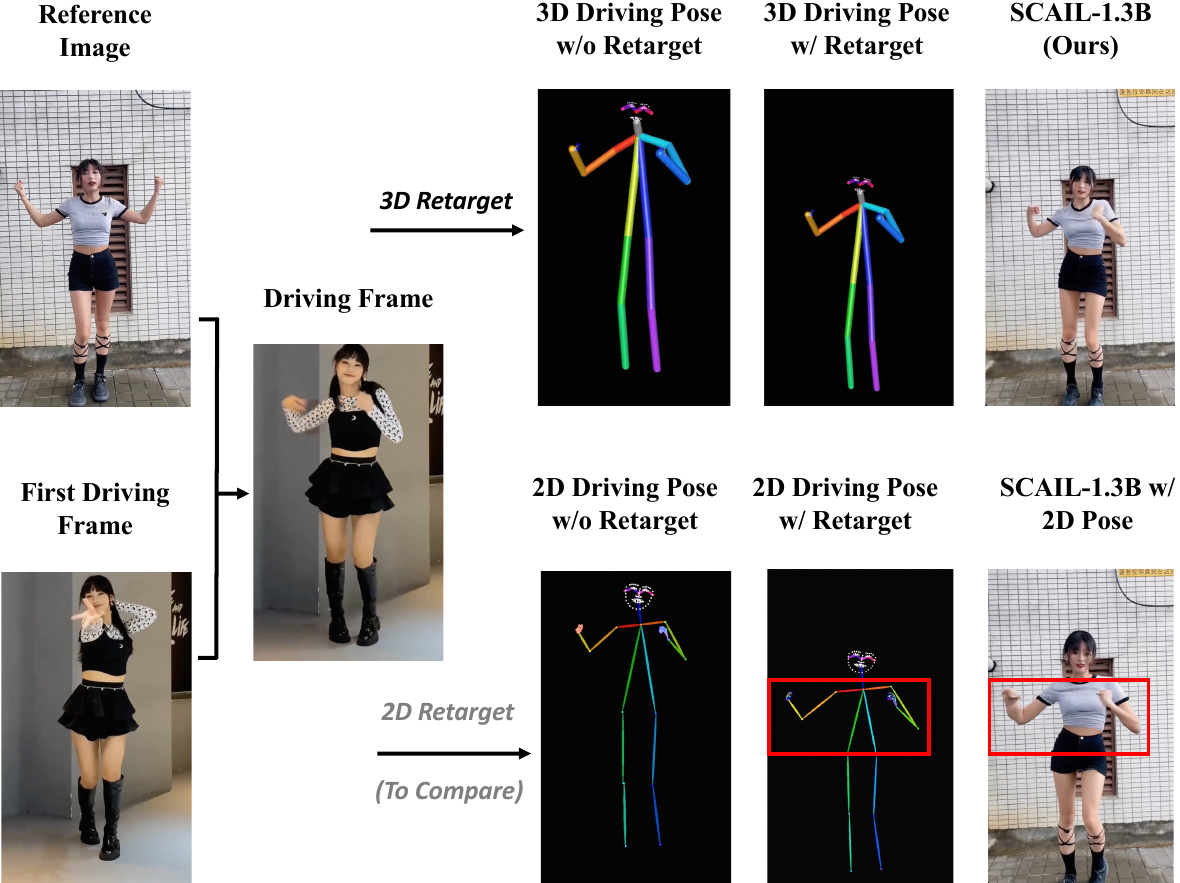}
    \caption{Ablation studies on pose retargeting. 2D Retarget are visualized for comparison. Regions where the body proportions deviate from the reference image are boxed.}
    \label{fig:ablationret}
\end{figure}

\noindent\textbf{Ablation on 3D Retarget.}
Figure~\ref{fig:ablationret} shows the comparison of the driving pose with and without 3D Retarget. 3D Retarget helps transfer the motion to the person without introducing position change. Compared to 2D Retarget, 3D Retarget keeps the original motion without introducing limb length distortion. Note that in our \textbf{Studio-Bench}, we only include cases where 2D Retarget works well for a fair comparison of the the model itself's performance against other baselines. In wild scenarios, however, 3D information and camera parameters can help create highly robust retarget rules that are suitable for production-level use.

\noindent\textbf{Ablation on 3D Augmentation.}
3D augmentation is central to our method's ability to adapt to different characters. We conducted experiments to evaluate the impact of using 3D augmentation in the context of self-driven animation. The results indicated that even with a high augmentation ratio, there was no significant difference in the metrics compared to when augmentation was not used. This is because 3D augmentation effectively preserves motion information by only altering figure shape and maintaining the temporal motion semantics.

\begin{figure}[t]
    \centering
    \includegraphics[width=\linewidth]{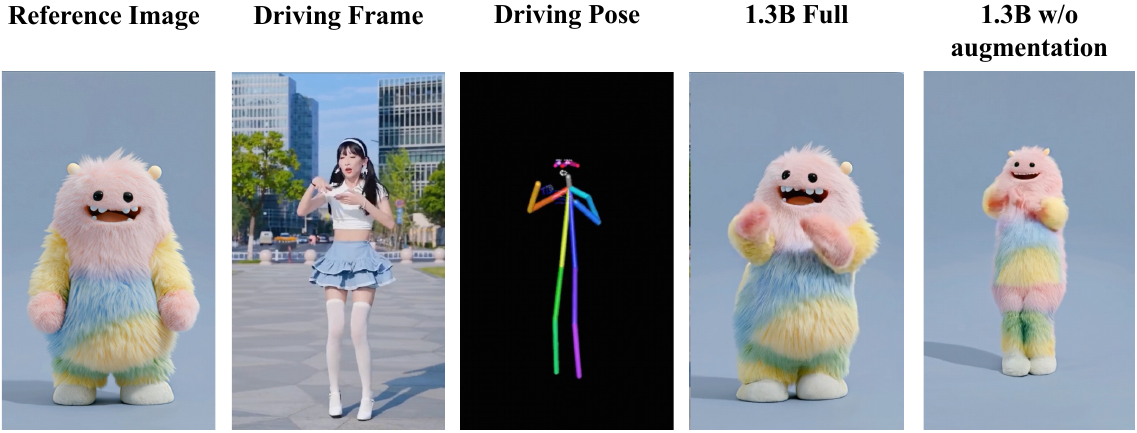}
    \caption{Qualitative results of ablation on 3D Augmentation.}
    \label{fig:ablationaug}
    \vspace{-0.2cm}
\end{figure}

In the case of cross-driven animation, 3D augmentation significantly mitigated identity leakage, particularly notable in characters with substantial body shape differences, as illustrated in Figure~\ref{fig:ablationaug}. To quantify this improvement, we conduct a user study comparing two groups: with and without 3D augmentation. The results reveal that 3D augmentation clearly enhances the metric of \textit{Physical Consistency} and \textit{Identity Similarity}, endowing our model with better generalization capabilities when handling in-the-wild characters.

\begin{figure}[t]
    \centering
    \includegraphics[width=\linewidth]{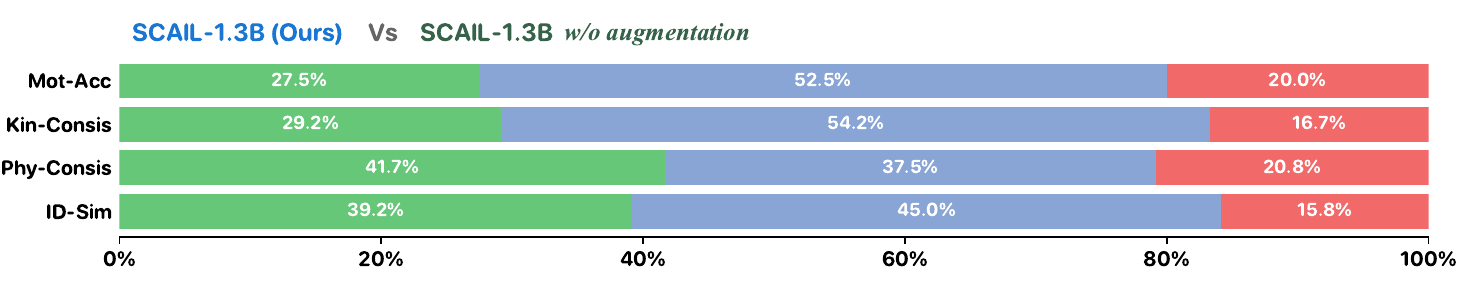}
    \caption{User study results of ablation on 3D Augmentation.}
    \label{fig:wintieloseabl}
\end{figure}

\section{Data Source}
Our training dataset is composed of three primary data sources:  
(1) samples retrieved from our internal base model training data and other downstream tasks,  
(2) a large collection of high-resolution dance videos from Bilibili and YouTube, and  
(3) additional sports videos such as gymnastics and figure skating. To ensure diversity, we maintain a certain proportion of stylized content, including 3D and 2D animations from source (1), as well as MMD and Live2D animations from source (2).

\section{Details on Studio-Bench}
\begin{figure}[t]
    \centering
    \includegraphics[width=0.85\linewidth]{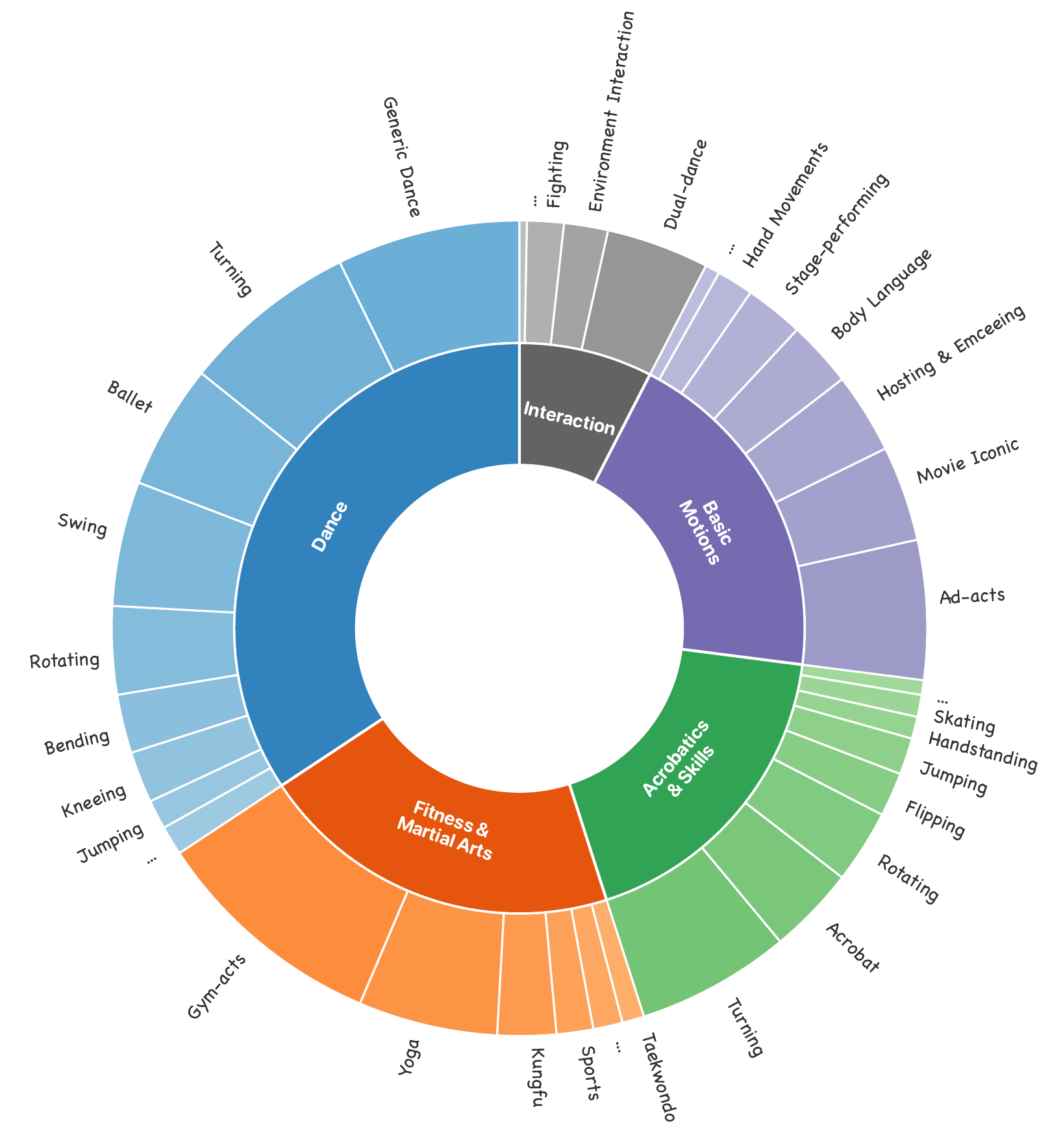}
    \caption{Visualization of the distribution of data annotations in \textbf{Studio-Bench}. We categorize and annotate videos based on motion types. The annotation of a single video can contain multiple tags, such as "turning" and "ballet".}
    \label{fig:bench_distribution}
    \vspace{-0.1cm}
\end{figure}

To comprehensively evaluate studio-grade scenarios, we curate an diverse set of motion sequences in our \textbf{Studio-Bench}, as illustrated in Figure~\ref{fig:bench_distribution}. The motion collection in our dataset emphasizes complex human-body configurations, covering a wide spectrum of challenging inputs, including dance, sports, martial-arts, acrobats and so on. In addition to isolated single-person motions, the test set also contains a small portion of interactions between the person and the environment, as well as several cases of multi-person interactions like dual dancing. We also include certain portion of fine-grained motions which are commonly featured in advertisement poses and iconic movie gestures, to evaluate the model's all-around capability.

For the construction of cross-driven cases, we intentionally select reference character images to cover diverse figure shapes and different facial characteristics. On top of these real human references, we additionally introduce approximately 40 non-real characters. Most of them originate from 3D animated productions, while others include 2D animated characters, plush toys, anime figurines, and various stylized representations.

\subsection{More Examples on Studio-Bench}
\begin{figure*}[htbp]
    \centering
    \setlength{\abovecaptionskip}{3pt}  
    \setlength{\belowcaptionskip}{0pt}
    \includegraphics[width=0.82\linewidth]{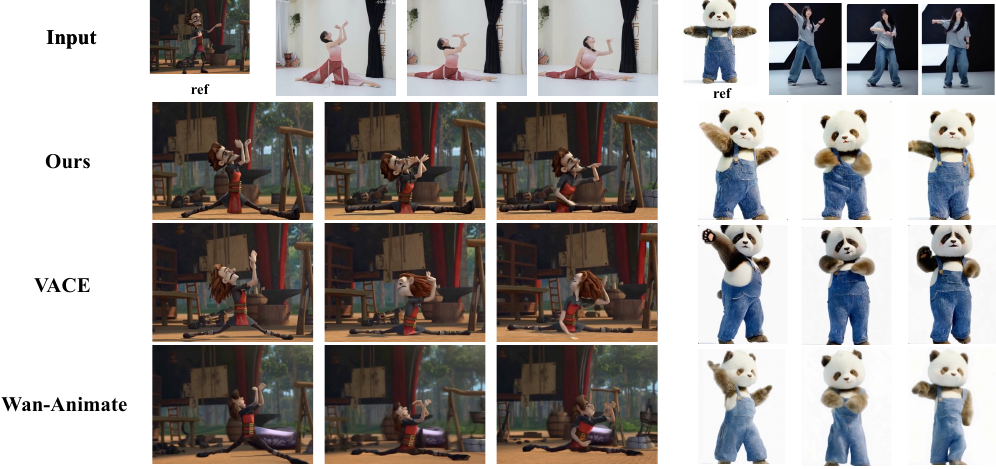}
    \caption{Comparison of model's ability to preserve body structure for non-standard character figures.}
    \label{fig:finalcompare2}
    \vspace{-0.135cm}
\end{figure*}

\begin{figure*}[htbp]
    \centering
    \setlength{\abovecaptionskip}{3pt}  
    \setlength{\belowcaptionskip}{0pt}
    \includegraphics[width=0.82\linewidth]{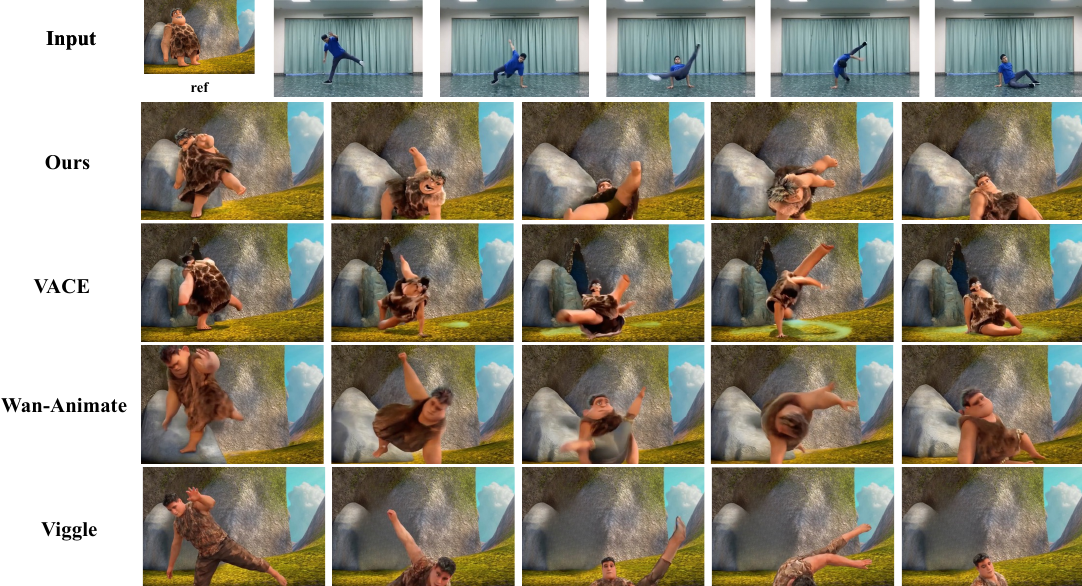}
    \caption{Visualization of our model's performance under both high-dynamic motion and non-standard character figures.}
    \label{fig:finalcompare3}
    \vspace{-0.135cm}
\end{figure*}

\begin{figure*}[htbp]
    \centering
    \setlength{\abovecaptionskip}{3pt}  
    \setlength{\belowcaptionskip}{0pt}
    \includegraphics[width=0.72\linewidth]{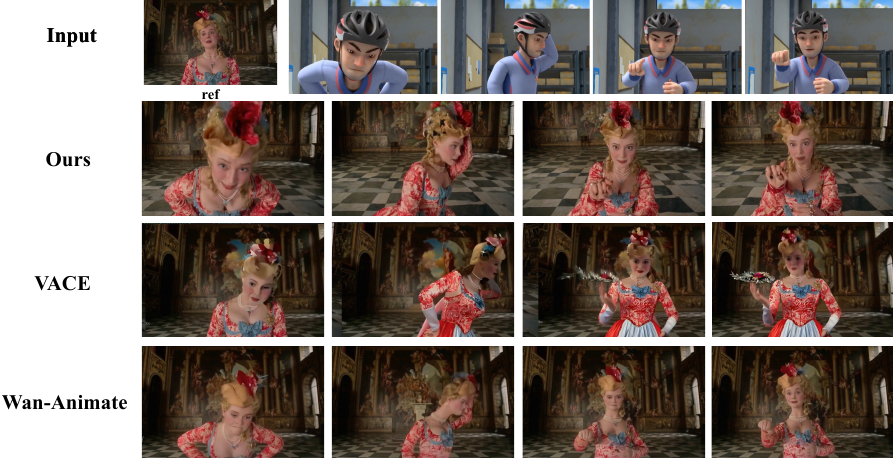}
    \caption{Visualization of our model's performance under \textit{reverse driving} settings.}
    \label{fig:finalcompare4}
    \vspace{-0.2cm}
\end{figure*}

Example from the main paper demonstrate that our model can handle both fine-grained motions and highly complex in-the-wild motions. Figure~\ref{fig:finalcompare2} and Figure~\ref{fig:finalcompare3} will provide more cases with nonstandard figures to show our model's generalization ability across a wide range of subjects and artistic styles.
Figure~\ref{fig:finalcompare2} demonstrates that SCAIL can produce accurate limb motions that respect the features of non-standard figures such as thin-limb anime characters. Furthermore, when the driving image is drastically different from the standard human figure (such as a plush toy with a very short body), SCAIL avoids the undesirable changes in body proportions that often plague baseline models. When the dual challenges of complex motions and non-standard character figures appear together in Figure~\ref{fig:finalcompare3}, the issues with baseline models become even more pronounced. Another point worth noting is the scenario of \textit{reverse driving}, where an anime character's motion is used to drive a real person. While most user inputs involve a real person driving another figure, this less common use case presents demand of making a real person mimic an anime posture. We found that previous methods produce strange proportions under such inputs in Figure~\ref{fig:finalcompare4} while our model is still capable of handling this task. These comparisons highlight the strong potential of our approach for studio-grade applications where character compatibility for diverse motion types are critical requirements.


\section{Discussion}

\subsection{Limitations and Future Work}
Although we have adopted a relatively effective multi-person pose extracting pipeline, the accuracy of multi-person pose estimation is still not as precise as that for single-person scenarios. While we are able to make the model more robust to inaccurate poses through tailored model architecture design and sufficient data, we still look forward to advances in the field of multi-person pose estimation to further improve the fidelity of motion replication.

Moreover, our \textbf{SCAIL} model currently relies on facial landmarks to achieve face control. Such a representation is inherently limited in fine-grained facial expression. As our work primarily targets addressing the challenges including instability and motion artifacts in studio-grade video generation, enhancing the expressiveness of facial control is left for future exploration. Specifically, future work will focus on improving the accuracy and fidelity of fine-grained details, such as hands and facial expressions, to further elevate the model's overall quality.

\subsection{Ethical Considerations}
Our approach is designed to produce studio-grade, high-fidelity character animation, enabling professional workflows in virtual production and cinematic pipelines. As our method advances character animation to a new level of realism and expressiveness, the potential for misuse, particularly in generating misleading or harmful digital content, becomes an important consideration. Despite these concerns, we believe that fully open-sourcing our model will bring substantial value to the community. By promoting transparency and broad accessibility, we aim to encourage a wide range of responsible, creative, and innovative works.

\end{document}